\documentclass[letterpaper, 10 pt, conference]{ieeeconf}  % Comment this line out if you need a4paper

\IEEEoverridecommandlockouts                              % This command is only needed if 
\usepackage{amssymb}  % assumes amsmath package installed
\usepackage[T1]{fontenc}
\usepackage{amsmath}
\usepackage{array}
\usepackage{multirow}
\usepackage{multicol}
\usepackage{graphicx}
\usepackage{hyperref}
\usepackage{threeparttable}
\usepackage{cite}

\title{\LARGE \bf PAPL-SLAM: Principal Axis-Anchored Monocular Point-Line SLAM}

\author{Guanghao Li{$^\dag$}, Yu Cao{$^\dag$}, Qi Chen{$^\dag$}, Yifan Yang, Jian Pu{$^\ast$}% <-this % stops a space
\thanks{$\ast$ Corresponding author}
\thanks{\dag \ These two authors.contribute equally to this work} 
\thanks{All authors are with Fudan University, Shanghai 200433, China (E-mails: \{ghli22, caoyu21, qichen21, yifyang23\}@m.fudan.edu.cn, jianpu@fudan.edu.cn).}
}

\begin{document}
\maketitle
\thispagestyle{empty}
\pagestyle{empty}

%%%%%%%%%%%%%%%%%%%%%%%%%%%%%%%%%%%%%%%%%%%%%%%%%%%%%%%%%%%%%%%%%%%%%%%%%%%%%%%%
\begin{abstract}
In point-line SLAM systems, the utilization of line structural information and the optimization of lines are two significant problems. The former is usually addressed through structural regularities, while the latter typically involves using minimal parameter representations of lines in optimization. However, separating these two steps leads to the loss of constraint information to each other. We anchor lines with similar directions to a principal axis and optimize them with $n+2$ parameters for $n$ lines, solving both problems together. Our method considers scene structural information, which can be easily extended to different world hypotheses while significantly reducing the number of line parameters to be optimized, enabling rapid and accurate mapping and tracking. To further enhance the system's robustness and avoid mismatch, we have modeled the line-axis probabilistic data association and provided the algorithm for axis creation, updating, and optimization. Additionally, considering that most real-world scenes conform to the Atlanta World hypothesis, we provide a structural line detection strategy based on vertical priors and vanishing points. Experimental results and ablation studies on various indoor and outdoor datasets demonstrate the effectiveness of our system. 
% Our code will be open-sourced upon paper acceptance.

% Traditional SLAM's reliance on point features often fails to utilize the structural information prevalent in scenes, leading to the emergence of point-line SLAM systems. However, these systems either do not explicitly harness structural information or use it too rigidly, imposing limitations on accuracy, speed, and application scope. We propose a monocular point-line SLAM system based on principal axis anchoring, which optimizes $n$ structural lines in the same direction using only $n+2$ variables by anchoring them to a same principal axis. This approach, accompanied by strategies for principal axis creation, deletion, optimization, and data association between structural lines and principal axes, fully leverages structural information in various directions without excessive rigidity. It significantly reduces the number of optimization parameters and enhances the system's speed, accuracy, and robustness. Furthermore, given that most structured scenes possess only one vertical orientation and many horizontal orientations (Atlanta world assumption), we also offer a structure line initialization and detection strategy based on vertical direction priors and vanishing points. Experimental results and ablation studies on multiple indoor and outdoor datasets demonstrate the robustness and efficacy of our principal axis-anchored monocular point-line SLAM system. Our code will be open sourced upon paper acceptance.

\end{abstract}

%%%%%%%%%%%%%%%%%%%%%%%%%%%%%%%%%%%%%%%%%%%%%%%%%%%%%%%%%%%%%%%%%%%%%%%%%%%%%%%%
\section{INTRODUCTION}

Simultaneous Localization and Mapping (SLAM), as a classical problem, has seen significant development over the past two decades. Among the numerous SLAM systems, monocular Vision SLAM (VSLAM) stands out as a classic system, widely adopted for its portability and low cost~\cite{davison2007monoslam}. However, most VSLAM systems~\cite{mur2015orb,mur2017orb,campos2021orb} are based on point features, which perform poorly in varying lighting conditions and low-texture areas (typically artificial environments). Consequently, recent years have seen a surge in research on point-line VSLAM systems~\cite{pumarola2017pl,gomez2019pl}, which have been showing useful~\cite{li2019leveraging} for the complex environment. The additional degree of freedom in line features compared to point features provides extra structural information but also introduces challenges in optimization. Therefore, two significant problems are utilizing the structural information provided by line features and efficiently optimizing it. 

The structural information of lines is evident as many line features exhibit the same (or similar) orientation, thus giving rise to SLAM systems based on hypotheses like the Manhattan World~\cite{coughlan1999manhattan}, Mixture of Manhattan World~\cite{straub2014mixture}, Atlanta World~\cite{schindler2004atlanta}, and Hong Kong World assumptions\cite{li2023hong}. These systems leverage specific patterns to impose constraints on system, yielding effective results. However, these systems do not integrate the structural information provided by assumptions with line optimization, limiting their applicability.

Line representation in SLAM can be categorized into over-parameterized and minimal parameter representations. Over-parameterized representations, such as representations using two endpoints, Edgelets~\cite{eade2009edge}, and Plücker coordinates~\cite{plucker1865xvii,sola2012impact}, are intuitive but require additional constraints during Bundle Adjustment (BA). This introduces extra computational overhead, numerical instability, and unsuitability for nonlinear optimization. On the other hand, minimal parameter representations like Orthogonal Representation~\cite{bartoli2005structure}, Cayley Representation~\cite{zhang2014structure}, and the No Singularities and Special Cases~\cite{roberts1988new} Method utilize four parameters to describe a line. These methods do not require additional constraints during optimization, but their capability for optimizing line orientation is limited~\cite{lim2022uv} and is less intuitive. Additionally, regardless of the method used, the optimization of lines is always slow.

Our work unifies the structural information of lines with line optimization, mitigating the challenges posed by each. We restore the line features to be optimized by utilizing the orientation of the anchored principal axis and the inverse depth of the midpoint of the observed line segment in the reference keyframe, which allows us to optimize \( n \) co-directional lines using only \( n+2 \) parameters. This approach enables easy extension to world hypotheses and achieves fast and accurate optimization. We have designed a corresponding management strategy for principal axes and an extended structural line detection strategy under the Atlanta World hypothesis. The main contributions of this paper are summarized as follows.

\begin{figure*}[t]
    \centering
    \includegraphics[width=\linewidth]{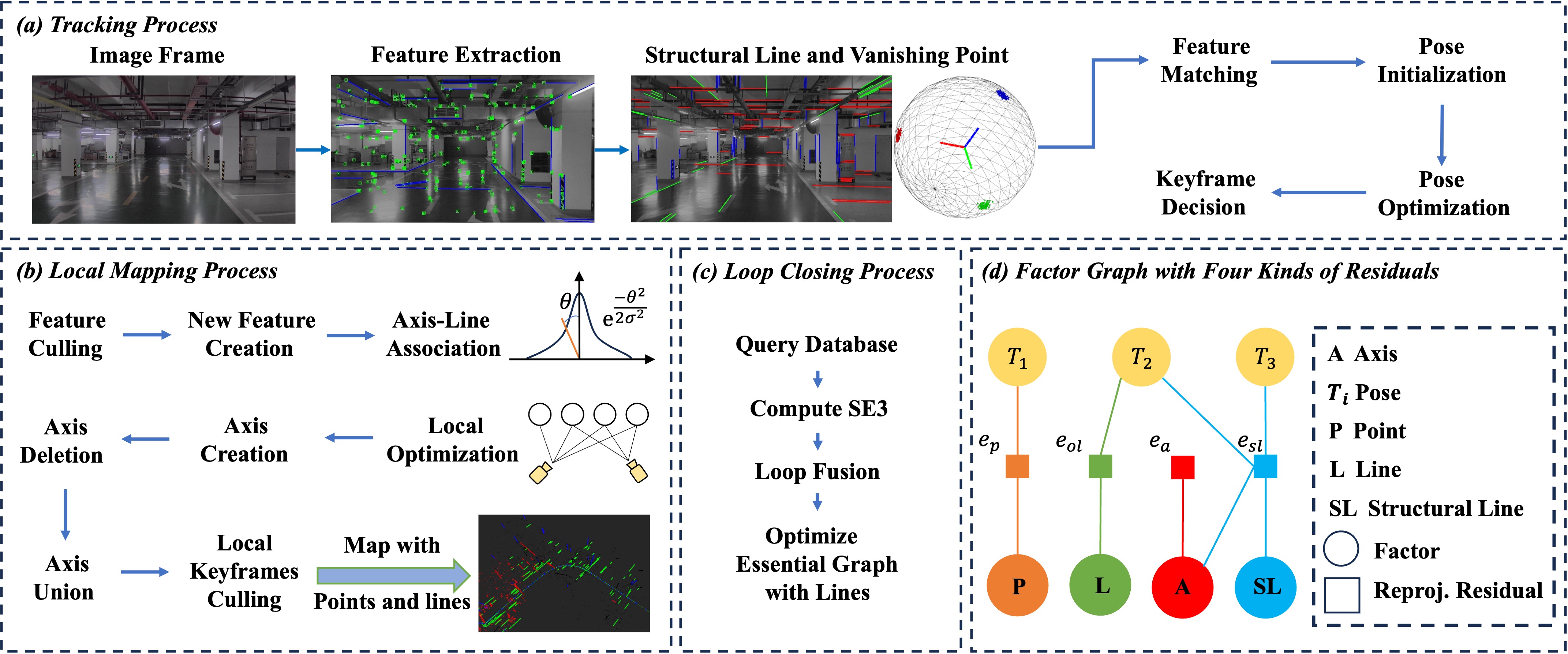}
    \caption{Overall of our system. Our system is based on VPL-SLAM~\cite{chen2024vpl} and ORB-SLAM2~\cite{mur2017orb}, consisting of tracking, local mapping, and loop closure threads. (a) shows the tracking thread workflow, where we detect structural lines in the scene and perform pose optimization. (b) shows the local mapping thread, where we perform principal axis management and local optimization. (c) show the loop closing process, using structural lines to optimize the essential graph. (d) shows our factor graph for BA, where we use a three-parameter optimization method based on principal axes in addition to traditional point and line feature optimization.}
    \label{fig:overall}
\end{figure*}

\begin{itemize}
    \item We utilize multiple principal axes to anchor structural line features, naturally and effectively incorporating the scene's structural information into the optimization process. 
    \item We model probabilistic data association between structural lines and principal axes and provide the strategies for principal axis creation, updating, and optimization, which enhances the system's robustness.
    \item We make an additional constraint for scenes commonly encountered under the Atlanta World hypothesis, specifically using the unique vertical direction in the map as a prior. We initialize related structural line features by utilizing vanishing points within this orientation.
    \item We conducted extensive experiments on our designed SLAM system, and the results have demonstrated its robustness and precision. These findings support the community, confirming the efficacy and reliability of our approach under various environments.
\end{itemize}

\section{Related Works}

\subsection{Point SLAM System}
Point features were widely adopted in most systems due to their simplicity and practicality. There were many methods for detecting point features, such as Harris corner~\cite{harris1988combined}, Shi-Tomasi corner~\cite{shi1994good}, SIFT~\cite{lowe2004distinctive}, FAST~\cite{rosten2006machine}, SURF~\cite{bay2008speeded}, Brief~\cite{calonder2010brief}, and ORB~\cite{rublee2011orb}. MonoSLAM~\cite{davison2007monoslam}, as the first real-time monocular VSLAM algorithm, used Shi-Tomasi corners~\cite{shi1994good} for tracking in the frontend and employed an Extended Kalman Filter (EKF) for optimization in the backend. PTAM~\cite{klein2007parallel} divided VSLAM into two threads: mapping and tracking. The tracking thread used FAST corners~\cite{rosten2006machine} for pose estimation, while the mapping thread replaced the EKF with a nonlinear optimization algorithm. LSD-SLAM~\cite{engel2014lsd} used a direct method by optimizing pixel intensities and performed loop closure with feature points. The ORB series~\cite{mur2015orb, mur2017orb, campos2021orb} adopted PTAM's dual-thread approach and introduced a loop closure detection thread. It used ORB features~\cite{rublee2011orb} for tracking and DBoW~\cite{galvez2012bags} for loop closure detection, making it one of the most influential SLAM systems today. Edge SLAM~\cite{maity2017edge} detected edge points from images and tracked those using optical flow. However, point features cannot be easily recognized and matched in areas with low textures and lighting variations. In such regions, fully utilizing the structural information of the scene can help improve localization and mapping.

\subsection{Point-Line SLAM System}
Some VSLAM systems~\cite{zhang2015building, pvribyl2016camera, vakhitov2016accurate, zuo2017robust, he2018pl, zhou2021dplvo, lim2021avoiding, zhang2023accurate, yan2024plpf, xu2024airslam, lee2019elaborate} used line features to add additional constraints to the scene, while others~\cite{xu2022leveraging, wei2022structural, shu2023structure} went further by using scene assumptions to enhance these constraints. For the former, \cite{smith2006real} used an EKF to jointly optimize point features and line features represented by two endpoints. PL-SLAM~\cite{pumarola2017pl} represented a 3D line with its two 3D endpoints and introduced a new reprojection error. EDPLVO~\cite{zhou2022edplvo} used the inverse depth of two endpoints for optimization. However, these systems did not effectively utilize the directional information of lines. For the latter, many systems were based on the Manhattan World hypothesis~\cite{zhou2015structslam,kim2017visual,kim2018low,li2018monocular,li2020structure,liu2020visual}, often optimizing the rotation matrix first and then the translation, but most were suited for indoor environments. Some systems were also based on the Atlanta World~\cite{li2019leveraging,zhou2024tracking,chen2024vpl} or Hong Kong World~\cite{li2023hong} hypotheses, providing a more general representation of structured scenes. UV-SLAM~\cite{lim2022uv} used structural regularities (vanishing point factor) without any constraints, making it more broadly applicable. However, the systems above do not effectively represent a 3D line with directional information in BA. Our system achieves this representation while also providing fast and accurate pose estimation.

\section{Method}
Given the known camera intrinsic matrix $\boldsymbol{K}$, the input to our system is a sequence of image frames $ \{ \boldsymbol{I}_t \}_{t=1}^{N} $, and the output consists of estimated poses $\boldsymbol{T}_{cw}$ and a map with point features $\boldsymbol{P}_n$ and line features $\boldsymbol{L}_n$. The pose $\boldsymbol{T}_{cw}$ is represented as an element of $SE(3)$ (the special Euclidean group), comprising a rotational component $\boldsymbol{R}_{cw} \in SO(3)$ (the special orthogonal group) and a translational component $\boldsymbol{t}_{cw} \in R^3 $. Fig.~\ref{fig:overall} illustrates the structure of our system. In sequence, we will introduce the system's point-line model, the principal axis management strategy, the line feature preprocessing strategy, and the joint optimization strategy of the system.

% in three dimension space
% (the special Euclidean group)
% (the special orthogonal group)

\subsection{Point-Line System Model}

In our system, we focus on the representation of lines, as the representation of points is referenced from ORB-SLAM2~\cite{mur2017orb}. We use a three-parameter representation to depict structural lines and an orthogonal representation for temporary non-structural lines during optimization. For their intuitive representation, we use end-points representation and Plücker coordinates in the non-optimization parts.
\begin{figure}[!t]
\centering
\includegraphics[width=\linewidth]{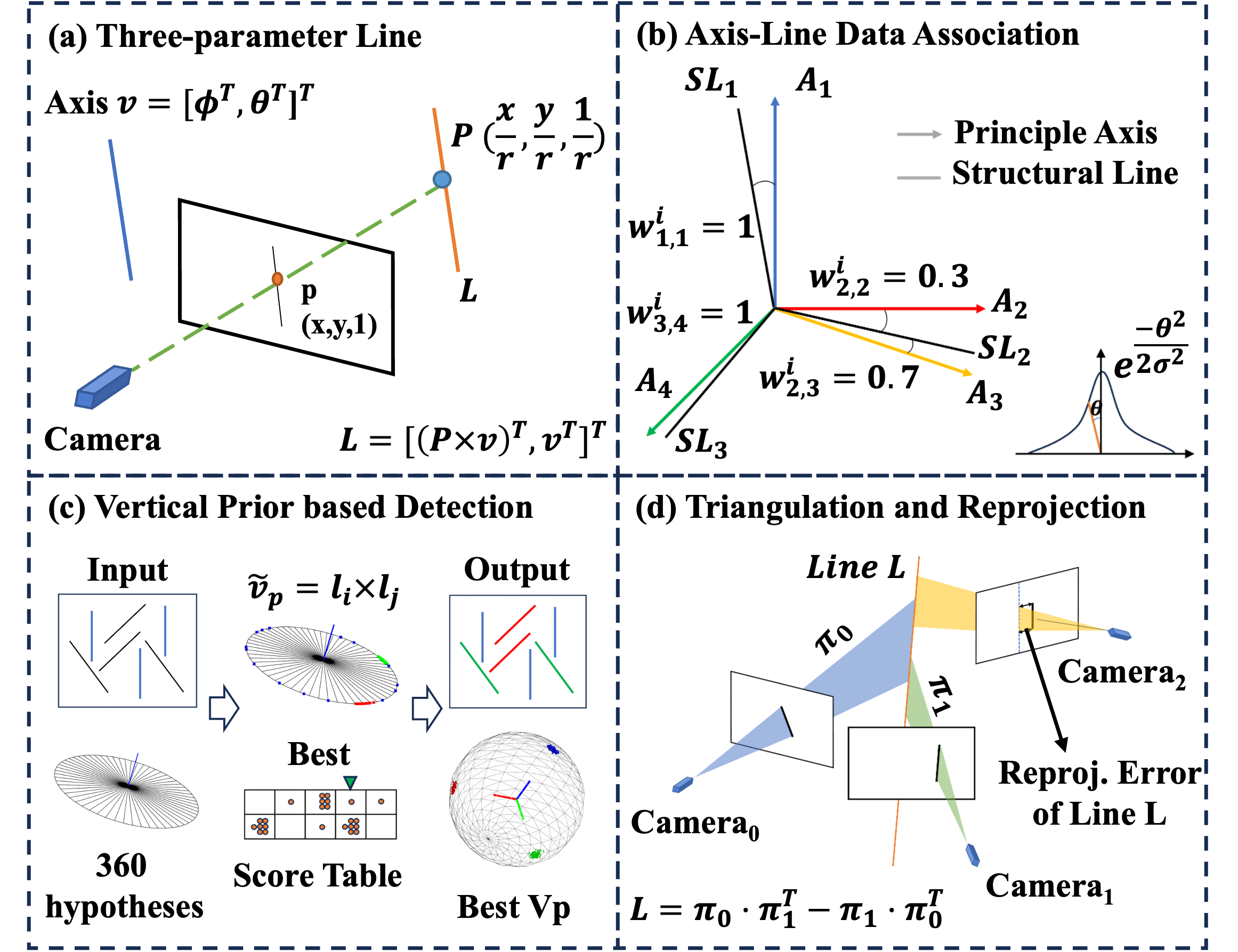}
\caption{Principal axis and line handling algorithms. (a) illustrates the definition of our three-parameter line representation. (b) defines our principal axis probabilistic association model. (c) shows the structural line and vanishing point detection strategy based on the vertical prior. (d) demonstrates the triangulation process of 3D lines and the reprojection error during optimization.}
\label{fig:lineParam}
\end{figure}

\subsubsection{Over-parameterized Representation} A 3D line $\boldsymbol L$ can be determined by two distinct points $\boldsymbol{P}_1, \boldsymbol{P}_2 $ in three-dimensional space, which is useful for visualization. Its plücker representation can be defined by the two points or by the direction vector $\boldsymbol{v}$ and a point $\boldsymbol{P}_3$ on the line:

\begin{equation}
    \boldsymbol{L} = 
    \begin{bmatrix} 
        \boldsymbol{n} \\ 
        \boldsymbol{v} 
    \end{bmatrix} = 
    \begin{bmatrix} 
        \boldsymbol{P}_1 \times \boldsymbol{P}_2 \\ 
        \boldsymbol{P}_2 - \boldsymbol{P}_1
    \end{bmatrix} = 
    \begin{bmatrix} 
        \boldsymbol{P}_3 \times \boldsymbol{v} \\ 
        \boldsymbol{v} 
    \end{bmatrix},
    \label{equ:plucker}
\end{equation}
where $\boldsymbol{n} \in R^3 $ is the normal vector of the plane determined by the line and the origin.

\subsubsection{Orthogonal Representation} 
We can use a three-dimensional rotation matrix $\boldsymbol U $ to represent the transformation of the coordinate system formed by the normal vector $\boldsymbol{n}$, direction vector $\boldsymbol{v}$, and their cross product in the Plücker representation relative to the original coordinate system, which determines the direction of the line. Additionally, we can use a two-dimensional rotation matrix $\boldsymbol{W}$ to represent the ratio of the magnitudes of the normal vector and the direction vector, which determines the distance of the line from the origin. By combining these two representations, we obtain the orthogonal representation $(\boldsymbol{U, W}) = (\boldsymbol{R}(\boldsymbol{\psi}), \boldsymbol{R}(\phi)) \in SO(3) \times SO(2)$ of the line~\cite{bartoli2005structure,he2018pl}.

\subsubsection{Three-parameter Representation.} For structural line $\boldsymbol{L}$, the inverse depth $r$ of the midpoint of the line segment in its reference keyframe and the direction $\boldsymbol v $ of its anchored principal axis $\boldsymbol v$ can determine its Plücker representation (Fig.~\ref{fig:lineParam} (a)):
\begin{equation}\label{eq_combined}
\boldsymbol{L} = \begin{bmatrix} \boldsymbol{P}^w \times \boldsymbol{v} \\ \boldsymbol{v} \end{bmatrix}, \quad \boldsymbol{\tilde{P}}^w = \boldsymbol{T}_{wc} \boldsymbol{\tilde{P}}^c, \quad \boldsymbol{P}^c = \boldsymbol{K}^{-1} \boldsymbol{\tilde{p}} / r.
\end{equation}
$\boldsymbol P^w$ and $\boldsymbol P^c$ represent the midpoint in the world coordinate system and the camera coordinate system, respectively, while $ \boldsymbol p $ represents the coordinates in the pixel coordinate system. The hat symbol $(\tilde{\cdot})$ above a letter indicates the corresponding homogeneous coordinates. Additionally, in order to avoid additional constraints during optimization, we use the latitude and longitude $(\phi,\theta)$ to represent the direction of the principal axis $\boldsymbol v$:

\begin{equation}
    \begin{cases}
    \phi =\arccos{(v_z/\sqrt{v_x^2+v_y^2+v_z^2})}\\
    \theta =\arctan{(v_x,v_y)}+\pi \\
    \end{cases}.
    \label{equ:direction}
\end{equation}

\subsection{Principal Axis Management}
% We introduce the processes of principal axis creation, data association, and update. For details on principal axis optimization, please refer to Sec.~\ref{sec:optimization}.

\subsubsection{Creation} 
When the number of unclassified structural lines in the scene reaches a certain threshold $\tau_n$, we identify potential principal axes among them. We model the process of finding principal axes as the problem of maximizing a density function of direction. Specifically, we use the mean shift~\cite{cheng1995mean} algorithm to cluster the direction vectors $ \boldsymbol d$ of all unclassified lines and identify the centers among them:
\begin{equation}
    \begin{array}{l}
    m(\boldsymbol d) = \frac{\sum_{\boldsymbol d_i \in \mathcal{D}_d} K\left(\angle(\boldsymbol{d_i}, \boldsymbol{d})\right) \boldsymbol d_i}{\sum_{\boldsymbol d_i \in \mathcal{D}_d} K\left(\angle(\boldsymbol{d_i}, \boldsymbol{d})\right)}, \\
    K\left(\angle(\boldsymbol{d_i}, \boldsymbol{d})\right) = e^{-c\left\|\angle(\boldsymbol{d_i}, \boldsymbol{d})\right\|^2}, 
    \end{array}
    \label{equ:meanshift}
\end{equation}
where $\mathcal{D}_d$ represents the set of other direction vectors within a certain angle $\tau_d$ of direction vector $\boldsymbol d$, and $\angle(\boldsymbol{d_i}, \boldsymbol{d})$ is the function used to calculate the angle between two vectors. $K(\cdot)$ is the Gaussian kernel function used to compute the weights based on different angles. We iteratively optimize Equ.~\ref{equ:meanshift} until convergence to obtain the potential new principal axis directions.

For potential new principal axes, we first compare them with existing axes. The new axis is discarded if the angle difference is smaller than $10^\circ$. We then calculate the average angle difference between the related unclassified structural lines of the potential axis and the potential axis to identify the optimal principal axis. Additionally, we calculate the ratio of the number of non-structural lines to the number of structural lines in the current local map scene, and only if this ratio exceeds $0.6$ will a new principal axis be created.

\subsubsection{Data Association}
The increase in principal axes and the drift in estimated poses can lead to incorrect line–principal axis association. To mitigate the impact of such incorrect associations and inspired by the approach to data association in semantic SLAM~\cite{bowman2017probabilistic}, we have probabilistically modeled the principal axis–structural line association. Given the estimated pose set $\boldsymbol{\mathcal{T}} \triangleq \{\boldsymbol T_{wc,t}\}_{t=1}^{N}$, principal axis set $\boldsymbol{\mathcal{A}} \triangleq \{\boldsymbol A_i\}_{m=1}^{M}$, and the set of structural lines $\boldsymbol{\mathcal{L}} \triangleq \{\boldsymbol L_{k}\}_{k=1}^{K}$ to be associated, the data association $\boldsymbol{\mathcal{D}} \triangleq \{ \alpha_k \}_{k=1}^{K}$, which indicates the association of structural line $\boldsymbol L_k$ with principal axis $\boldsymbol A_{\alpha_k}$, is estimated. With initial estimates $\boldsymbol{\mathcal{T}}^{i}$ and $\boldsymbol{\mathcal{A}}^{i}$, unlike previous hard associations, we calculate an optimal estimate by maximizing the expected measurement likelihood using the Expectation Maximization (EM) algorithm, which considers the density distribution of all possible $\boldsymbol{\mathcal{D}}$:
% This soft association method takes into account all potential scenarios
\begin{equation}
    \begin{aligned}
    & \underset{\mathcal{T}, \mathcal{A}}{\arg \max } \mathbb{E}_{\mathcal{D}}\left[\log p(\mathcal{L} \mid \mathcal{T}, \mathcal{A}, \mathcal{D}) \mid \mathcal{T}^i, \mathcal{A}^i, \mathcal{L}\right] \\
    & =\underset{\mathcal{T}, \mathcal{A}}{\arg \max } \sum_{\mathcal{D} \in \mathbb{D}} p_v\left(\mathcal{D} \mid \mathcal{T}^i, \mathcal{A}^i, \mathcal{L}\right) \log p(\mathcal{L} \mid \mathcal{T}, \mathcal{A}, \mathcal{D}) \\
    % & =\underset{\mathcal{T}, \mathcal{A}}{\arg \max } \sum_{\mathcal{D} \in \mathbb{D}} \sum_{k=1}^K p_v\left(\mathcal{D} \mid \mathcal{T}^i, \mathcal{A}^i, \mathcal{L}\right) \log p\left(\boldsymbol{L}_k \mid \boldsymbol{\mathcal{T}}, \boldsymbol A_{\alpha_k}\right) \\
    & =\underset{\mathcal{T}, \mathcal{A}}{\arg \max } \sum_{k=1}^K \sum_{j=1}^M w_{k j}^i \log p\left(\boldsymbol{L}_k \mid \boldsymbol{\mathcal{T}}, \boldsymbol A_{j}\right), \\
    & w_{k j}^i = \sum_{\mathcal{D}\in \boldsymbol{\mathbb{D}}(k,j)} \frac{p_v\left(\mathcal{L} \mid \mathcal{T}^{i}, \mathcal{A}^{i}, \mathcal{D}\right)}{\sum_{\mathcal{D} \in \mathbb{D}} p_v\left(\mathcal{L} \mid \mathcal{T}^{i}, \mathcal{A}^{i}, \mathcal{D}\right)},
    \end{aligned}
\end{equation}
where $\mathbb{D}$ is the space of all possible values of $\boldsymbol{\mathcal{D}}$, and $w_{k j}^i$ represents the combined possibility of the set $\boldsymbol{\mathbb{D}} (k,j)$, which contains all possible associations where structural line $\boldsymbol{L}_k$ is associated with principal axis $\boldsymbol{A}_j$. Besides, $p_v\left(\mathcal{L} \mid \mathcal{T}^{i}, \mathcal{A}^{i}, \mathcal{D}\right)$ is different from $p\left(\mathcal{L} \mid \mathcal{T}, \mathcal{A}, \mathcal{D}\right)$, because the former only considers the directional information for line-axis association without considering positional information.
%\triangleq \sum_{\boldsymbol{\mathcal{D}}\in \boldsymbol{\mathbb{D}}(k,j)}   p_v\left(\mathcal{D} \mid \mathcal{T}^i, \mathcal{A}^i, \mathcal{L}\right)

% $p_v\left(\boldsymbol{L}_k \mid \boldsymbol{\mathcal{T}}, \boldsymbol A_{j}\right)$
Fig.~\ref{fig:lineParam} (b) shows the data association process. To reduce computational load, we first calculate the angle between the direction vector of each structural line $\boldsymbol{L}_k$ and the direction of the principal axis $\boldsymbol{A}_m$. We use a relatively loose threshold to filter out lines with large angles, defining their weight $w_{k,m}$ as 0. For the remaining lines, we assume that the probability that a line belongs to an axis follows a normal distribution with a mean of $0^\circ$. The input is the average angle between the vanishing point direction and the principal axis across all frames where the structural line $\boldsymbol{L}_k$ is observed. 
% $\boldsymbol{p}_v(\cdot)$ 

\subsubsection{Update}
We calculate the angle difference between the principal axis $\boldsymbol{A}_m$ before and after the BA. If the angle difference exceeds $\tau_{diff}$, the direction of the principal axis after the BA is adopted. It will be deleted if a principal axis undergoes significant changes multiple times. For adjacent principal axes, the latter-detected axis will be deleted.
% Based on the number $N$ of structural lines associated with the principal axis (for each structural line $\boldsymbol{L}_k$, the principal axis with the highest $w_{k,:}$ is considered its associated principal axis), different thresholds $\tau_{diff}$ are set (Equ.~\ref{equ:angleDiff}).
% For principal axes that show minimal updates over multiple optimizations, they are considered close to the true scene value and will be fixed and no longer optimized.
% \begin{equation}
%     \tau_{diff}=\left\{\begin{aligned}
%     10^{\circ} & (N<=20) \\
%     5^{\circ} & (20<N<50) \\
%     1^{\circ} & (N>50)
%     \end{aligned}.\right.
%     \label{equ:angleDiff}
% \end{equation}

\subsection{Structural Preprocess}
We use EDLines~\cite{akinlar2011edlines} to detect line segments, as it can detect more structural lines~\cite{chen2024vpl} with lower noise compared to LSD~\cite{von2008lsd}. After detecting the line segments, we follow the approach outlined in \cite{chen2024vpl} for line segment fusion and matching. Finally, we extract vanishing points and structural lines from the images similar to the method described in \cite{lu20172}.

\subsubsection{AW line Preprocess}
Given that most structured scenes in everyday life conform to the Atlanta World hypothesis, we additionally provide a structural line and vanishing point detection algorithm based on vertical priors (Fig.~\ref{fig:lineParam} (c)). Specifically, we modify the vanishing point detection algorithm proposed in \cite{lu20172}. In the early stages of system operation, we extract vertical lines from the images at the system's frontend following the method outlined in \cite{chen2024vpl}. The system begins extracting horizontal structural lines once the vertical principal axis \( \boldsymbol{d}_v \) is initialized. 

We extract two primary horizontal vanishing point directions in each frame for stability. The two directions should lie on the plane \( \boldsymbol \pi_v \) perpendicular to  \( \boldsymbol{d}_v \). Assuming \( \boldsymbol{d}_v=[\sin a \sin b, \sin a \cos b, \cos a]^\top \), we can derive a direction vector \( \boldsymbol{d}_h=[\cos b, -\sin b, 0]^\top \) perpendicular to \( \boldsymbol{d}_v \). By rotating \( \boldsymbol{d}_h \) around \( \boldsymbol{d}_v \) at different angles \( \theta_i \) with step $1^\circ$, we can generate $\boldsymbol{d}_{h,\theta_i}$ and all proposals \( \boldsymbol h_i=\{{\boldsymbol{d}_v, \boldsymbol{d}_{h,\theta_i}, \boldsymbol{d}_v \times \boldsymbol{d}_{h,\theta_i}}\}_{i=1}^{360} \). Subsequently, the proposals \( \{\boldsymbol h_i\}_{i=1}^{N} \) are scored as \cite{lu20172} does, and the optimal proposal is selected. Through the vertical axis prior, the number of proposals is significantly reduced from the original 37,800 proposals~\cite{lu20172} down to 360 proposals.

Further optimization is necessary because the number of samples limits the accuracy of the initial proposal. We use the distance from the lines to the two optimal vanishing points to find the corresponding line sets \( \boldsymbol{\mathcal{C}}_0, \boldsymbol{\mathcal{C}}_1 \). For each set $\boldsymbol{\mathcal{C}}$, We construct the matrix \( \boldsymbol{M} \), where \( \boldsymbol{M}_i = \tilde{\boldsymbol{s}}_i \times \tilde{\boldsymbol{e}}_i \) represents the parametric equation of the line segment in the row $i$, with \( \bar{\boldsymbol{s}}_i \) and \( \bar{\boldsymbol{e}}_i \) being the homogeneous coordinates of the observed line segment endpoints. The solution to the matrix equation \( \boldsymbol{M}\boldsymbol{x} = 0 \) represents the intersection point of the line segments in the set. By performing Singular Value Decomposition (SVD) on the matrix \( \boldsymbol{M} \), we obtain the final vanishing point \( vp_{\text{fine}} \). After optimization, the refined vanishing point reclassifies the structural line segments. 

\subsection{Optimization}
\label{sec:optimization}
We focus on optimizing lines in the map while optimizing points can be referenced from ORB-SLAM2~\cite{mur2017orb}. During the initial optimization of map lines not yet associated with a principal axis, we treat their vanishing point direction in the corresponding keyframe as a temporary principal axis and optimize using our three-parameter representation. In subsequent optimizations, we employ the orthogonal representation until the line is associated with at least one principal axis. At this point, we revert to three-parameter optimization. This three-stage optimization approach, particularly the first stage, helps mitigate the difficulty of optimizing directions using the orthogonal representation~\cite{lim2022uv} and lays a solid foundation for correct subsequent associations.

\subsubsection{Reprojection Error}
Fig.\ref{fig:lineParam}(d) shows the process of 3D line triangulation and its reprojection error in the image. For a 3D line $\boldsymbol {L_w}$,  we first translate it from the world coordinate to the current frame coordinate:
\begin{equation}
\boldsymbol{L_c} = \boldsymbol{T}_{cw} \boldsymbol{L}_w =
\begin{bmatrix}
\boldsymbol{R}_{cw} & {[\boldsymbol{t}_{cw}]}_{\times} \boldsymbol{R}_{cw} \\
\boldsymbol{0} & \boldsymbol{R}_{cw} \\
\end{bmatrix}
\boldsymbol{L_w} =
\begin{bmatrix}
\boldsymbol{n}_c \\
\boldsymbol{v}_c \\
\end{bmatrix},
\end{equation}
where $[\cdot]_{\times}$ is the symbol for the antisymmetric matrix transformation for a vector. Then, we project the line $\boldsymbol{L}_c$ onto the pixel plane and calculate the reprojection error as the distance from the endpoints ${\boldsymbol{p}_1, \boldsymbol{p}_2}$ of the line segment to the projected line:
\begin{equation}\label{eq7}
\boldsymbol {l^{'}}=\boldsymbol {K_{L}} \boldsymbol {n_{c}} =
\begin{bmatrix}
f_y & 0 & 0 \\
0 & f_x & 0 \\
-f_yc_x & -f_xc_y & f_xf_y \\ 
\end{bmatrix} \boldsymbol n_{c} =
\begin{bmatrix}
l_1 \\
l_2 \\
l_3 \\
\end{bmatrix}
{\in R^3},
\end{equation}
\begin{equation}
\label{eq8}
\boldsymbol e_l={\left[\frac{{\boldsymbol {p_{1}}}^{T} \boldsymbol {l^{'}}}{\sqrt{{l_1}^2+{l_2}^2}},\frac{{\boldsymbol p_2}^{T} \boldsymbol l^{'}}{\sqrt{{l_{1}}^{2}+{l_{2}}^{2}}}\right]}^{T}.
\end{equation}

\begin{figure}[!t]
\centering
\includegraphics[width=\linewidth]{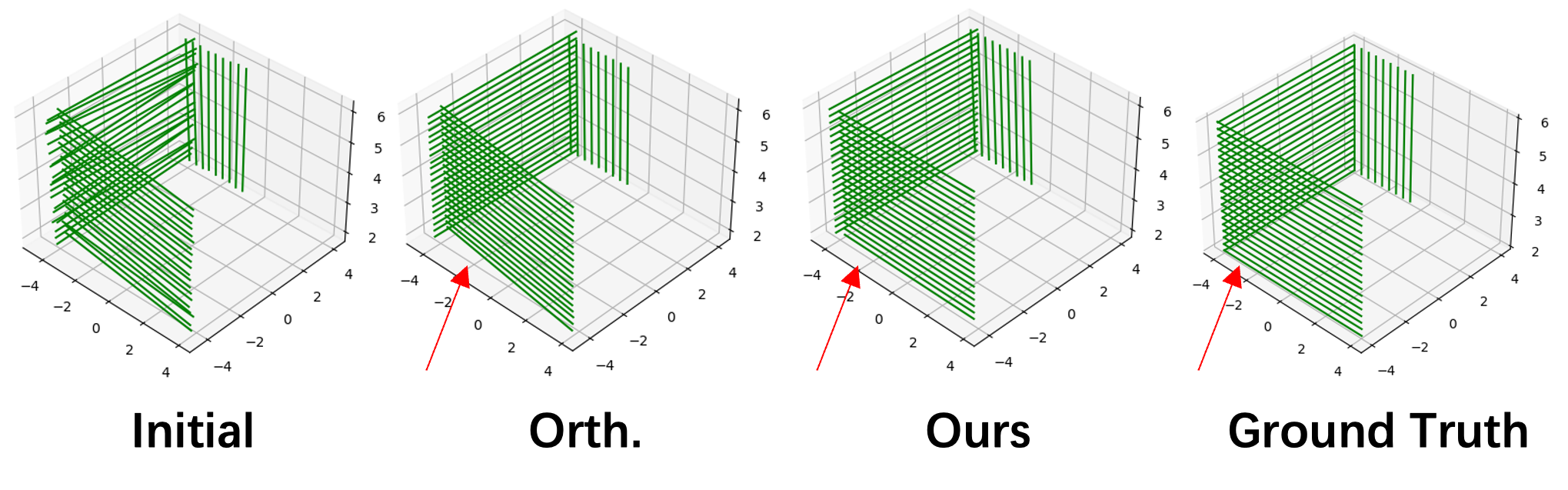}
\caption{Qualitative comparison between our representation and the orthogonal representation before and after optimization, under given pose and observation perturbations.}
\label{fig:compareline}
\end{figure}

\begin{table}[t]
\centering
\caption{Results on Synthetic Data}
\tabcolsep=0.1cm
\scalebox{0.2}
\small
\begin{tabular}{c|ccc|ccc|ccc}
\hline
\multirow{2}{*}{Metric} & \multicolumn{3}{c|}{Fix Pose} & \multicolumn{3}{c|}{Small Pose Noise} & \multicolumn{3}{c}{Large Pose Noise}\\
             & 2-p & 4-p & 3-p & 2-p & 4-p & 3-p & 2-p & 4-p & 3-p \\ 
\hline
Time       &0.956 &1.097 &\textbf{0.251} &5.977 &7.359 &\textbf{4.686}  &6.934 &8.767 &\textbf{5.100} \\
$\text{Error}_{l}$        &0.014 &\textbf{0.006} &0.009 &0.596 &0.942 &\textbf{0.314}  &6.577 &5.144 &\textbf{3.643} \\
Trans.      &\textbf{0} &\textbf{0} &\textbf{0} &0.107 &0.109 &\textbf{0.072}  &0.787 &0.775 &\textbf{0.747} \\ 
\hline
\end{tabular}
\label{tab:compression_time}
\begin{tablenotes}
    \footnotesize
    \item[-] $\text{Error}_l$ represents the Plücker representation residual norm, Trans. is the ATE RMSE of the pose, and the unit for Time is seconds.
    \end{tablenotes}
\end{table}

\subsubsection{Optimization Function}
The following are four different types of optimization errors (Fig.\ref{fig:overall}(d)) in our factor graph: 
\begin{equation}
    \begin{aligned}
    &\begin{aligned}
    \mathcal{T}^{i+1}, \mathcal{A}^{i+1}= & \underset{\mathcal{T}, \mathcal{A}}{\arg \min } \sum_{k=1}^K \sum_{j=1}^M w_{k j}^i  \boldsymbol{e}_{sl}^T \boldsymbol{\Omega}_{sl} \boldsymbol{e}_{sl} + \sum \boldsymbol{e}_{p}^T \boldsymbol{\Omega}_{p} \boldsymbol{e}_{p} \\
    & + \sum \boldsymbol{e}_{ol}^T \boldsymbol{\Omega}_{ol} \boldsymbol{e}_{ol} + \sum \boldsymbol{e}_{a}^T \boldsymbol{\Omega}_{a} \boldsymbol{e}_{a},
    \end{aligned}\\
    \end{aligned}
\end{equation}
where \( e_{sl} \) is the structural line three-parameter optimization error considering data association or temporary vanishing points, \( e_{ol} \) is the orthogonal representation error for structural lines not bound to a principal axis, $\boldsymbol{e}_a$ is the axis error that prevents excessive changes in the direction of the principal axis, and \( e_{p} \) is the point reprojection error. It is worth to notice that $\boldsymbol{e}_{sl}$ corresponds to the probability $\log p\left(\boldsymbol{L}_k \mid \boldsymbol{\mathcal{T}}, \boldsymbol A_{j}\right)$ with similar form in \cite{bowman2017probabilistic}, which means the reprojection error that projecting 3D line $\boldsymbol{L}_k$ associated with axis $\boldsymbol{A}_j$ onto all of the frames that observed line $\boldsymbol{L}_k$. 

% \subsubsection{Jacobian}
% We briefly shows the chain rule of the re-projection error $e_l $ with respect to the three-parameter line $\left[\phi, \theta\right]^{\top}, r$ . The computation results for pose variables follow the same approach as described in \cite{zuo2017robust}.
% \begin{equation}
% \centering
% \begin{aligned}
% \frac{\partial e_l}{\partial\left[\phi, \theta\right]^{\top}}&=\frac{\partial e_l}{\partial \boldsymbol{v}}  \frac{\partial \boldsymbol{v}}{\partial[\phi, \theta]^{\top}}, 
% \frac{\partial e_l}{\partial r}=\frac{\partial e_l}{\partial \boldsymbol{L}_w} \frac{\partial \boldsymbol{L}_w}{\partial r}, \\
% % \frac{\partial e_l}{\partial \delta \boldsymbol{\xi}_{s w}}&=\frac{\partial e_l}{\partial \boldsymbol{P}_{w}} \frac{\partial \boldsymbol{P}_w}{\partial \delta \boldsymbol{\xi}_{s w}}, 
% % \frac{\partial e_l}{\partial \delta \boldsymbol{\xi}_{cw}}=\frac{\partial e_l}{\partial \boldsymbol{L}_c} \frac{\partial \boldsymbol{L}_c}{\partial \delta \boldsymbol{\xi}_{cw}}, \\
%  \frac{\partial \boldsymbol{v}}{\partial[\phi, \theta]^{\top}}&=\left[\begin{array}{cc}
% \cos \phi \sin \theta & \sin \phi \cos \theta \\
% \cos \phi \cos \theta & -\sin \phi \sin \theta \\
% -\sin \phi & 0
% \end{array}\right]_{3 \times 2}, \\
% \frac{\partial \boldsymbol{L}_w}{\partial r}&=\left[\begin{array}{c}\frac{[\boldsymbol{v}]_{\times} \boldsymbol{R}_{ws} \boldsymbol{P}_s}{r} \\
% \boldsymbol{0}^{3 \times 1}\end{array}\right]_{6 \times 1}.
% \end{aligned}
% \end{equation}

\section{Experiments}
% & 2-param. & 4-param. & 3-param.
\begin{figure}[!t]
\centering
\includegraphics[width=\linewidth]{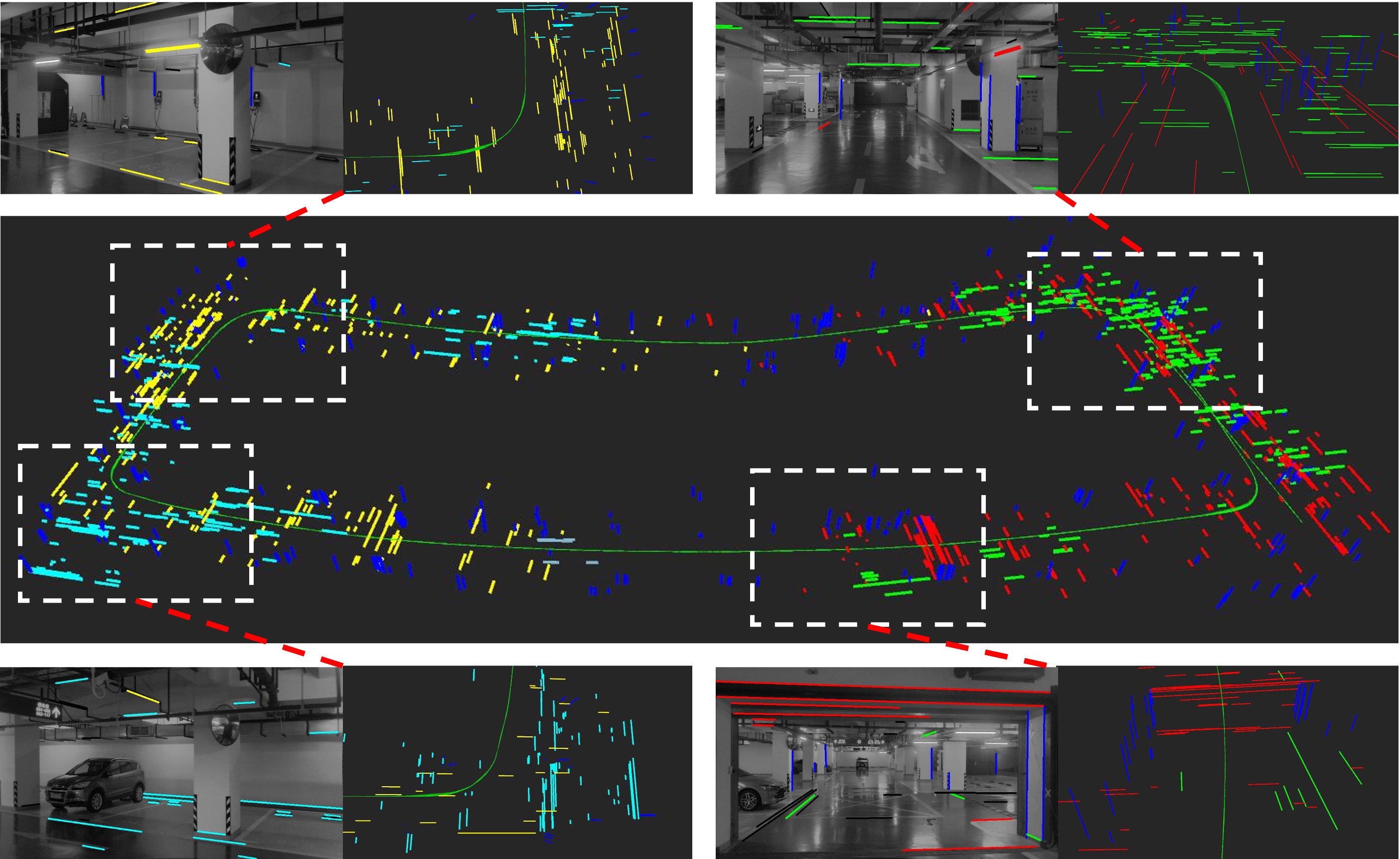}
\caption{The qualitative mapping results on the garage dataset show lines in different colors representing different principal axes. Using line features and various colors, our map visually and intuitively illustrates the corresponding scene.}
\label{fig:indoorVis}
\end{figure}
\subsection{Implementation Details}

We run our system on a desktop PC with an Intel Core i7-12700 CPU. To evaluate the robustness of our system, we test it in complex outdoor datasets KITTI~\cite{geiger2013vision} and Campus~\cite{chen2024vpl} and the challenging underground parking dataset (with glare and blur) BeVIS~\cite{shao2023slam} and Garage~\cite{chen2024vpl}. Moreover, we generated a toy dataset to compare the performance of principal axis-anchored optimization with other line feature optimization methods. 

To assess the tracking and mapping capabilities of our system, we compared it with open-sourced point-based SLAM systems ORB-SLAM3~\cite{campos2021orb}, LDSO~\cite{gao2018ldso}, and open-sourced point-line(-plane) based SLAM systems Structure-SLAM~\cite{li2020structure} (For outdoor environment, we use its point-line mode without the indoor structural constraint), Structure PLP-SLAM~\cite{shu2023structure}, and VPL-SLAM~\cite{chen2024vpl}. We used absolute trajectory error (ATE) root mean square error (RMSE) as the metric for tracking evaluation. All metrics were evaluated as the average of 10 test runs for fairness.

\begin{table*}[t]
    \centering
    \caption{ATE RMSE (Unit: M) Results in BeVIS, Garage and Campus Daatsets}
    \begin{tabular}{c|ccccc|cccc|ccccc}
    \hline
    \tabcolsep=0.2cm
    \multirow{2}{*}{Method} & \multicolumn{5}{c|}{BeVIS} & \multicolumn{4}{c|}{Garage} & \multicolumn{5}{c}{Campus} \\
    & $00$ & $01$ & $02$ & $03$ & Avg. & $00$ & $01$ & $02$ & Avg. & $00$ & $01$ & $02$ & $03$ & Avg. \\ 
    \hline
    LDSO~\cite{gao2018ldso} & $0.53^*$ & $1.49^*$ & $0.47^*$ & 0.18 & 0.67 &1.43 & 1.20 & 1.02 & 1.22 & 0.67 & 1.74 & 2.21 & 5.10 & 2.43 \\
    ORB-SLAM3~\cite{campos2021orb} & 0.41 & \underline{0.04} & \underline{0.29} & 0.26 & \underline{0.25} &1.75 & 2.08 & 1.14 & 1.66 & 1.18 & 1.61 & 2.56 & 4.37 & 2.43 \\
    Structure-SLAM~\cite{li2020structure} & \underline{0.33} & - & 12.64 & \underline{0.13} & - & - & - & - & - & 1.32 & 1.88 & 3.84 & - & - \\
    Structure PLP-SLAM~\cite{shu2023structure} & $1.89^*$  & \textbf{0.03} & - & 0.18 & - &3.19  & 6.73 & 1.13 & 3.68 & 2.40 & 5.67 & 6.37 & 20.77 & 8.80 \\
    VPL-SLAM~\cite{chen2024vpl}  & 0.37 & 0.16 & 0.41 & 0.16 & 0.28 & \underline{0.85} & \underline{1.05} & \underline{1.01} & \underline{0.97} & \underline{0.65} & \underline{0.87} & \underline{2.12} & \underline{3.62} & \underline{1.82} \\ 
    Ours & \textbf{0.09} & \underline{0.04} & \textbf{0.21} & \textbf{0.09} & \textbf{0.11} & \textbf{0.69} & \textbf{0.75} & \textbf{0.91} & \textbf{0.78} & \textbf{0.61} & \textbf{0.77} & \textbf{1.78} & \textbf{3.02} & \textbf{1.54} \\ 
    \hline
    \end{tabular}
    \label{tab:under}
    \begin{tablenotes}
    \item[-] *: The system crashed during operation, but there was some pose information recorded before the crash.
    \item[-] -: The systems tend to get lost during tracking because of visual degradation. 
    \end{tablenotes}
\end{table*}

\begin{table*}[t]
    \centering
    \begin{minipage}{0.64\textwidth}
    \begin{center}
        \caption{ATE RMSE (Unit: M) Results in KITTI Odometry dataset}
        \label{tab:kitti}
        \tabcolsep=0.14cm
        \begin{tabular}{c|ccccc|cccccc}
        \hline 
        \multirow{2}{*}{Method} & \multicolumn{5}{c|}{Structured} & \multicolumn{6}{c}{Semi-structured} \\ 
        & $00$ & $05$ & $06$ & $07$ & $08$ & $01$ & $02$ & $03$ & $04$ & $09$ & $10$ \\ 
        \hline\
        LDSO~\cite{gao2018ldso} & 9.32 & 5.10 & 13.55 & 2.96 & 129.02 & \textbf{11.68} & 31.98 & 2.85 & 1.22 & 21.64 & 17.36 \\
        ORB3~\cite{campos2021orb} & 8.07 & 6.71 & 15.19 & 2.89 & \underline{55.97} & - & 25.38 & 1.05 & 1.25 & \underline{8.04} & 8.76 \\
        Struct.~\cite{li2020structure}  & 6.62 & 12.62 & 23.67 & 3.36 & 104.92 & - & 23.62 & 2.68 & 1.22 & 13.78 & \textbf{7.52}  \\
        PLP~\cite{shu2023structure} & 7.16 & 9.67 & 20.42 & 4.91 & 66.99 & - & 34.23 & 7.21 & \textbf{0.47} & 24.98 & 11.31\\
        VPL~\cite{chen2024vpl} & \underline{6.04} & \underline{4.64} & \textbf{11.01} & \underline{1.64} & 66.05 & - & \underline{23.23} & \underline{0.81} & 0.75 & 8.22 & \underline{8.49} \\ 
        Ours & \textbf{5.62} & \textbf{4.51} & \underline{12.47} & \textbf{1.52} & \textbf{50.87}  & - & \textbf{21.36} & \textbf{0.77} & \underline{0.72} & \textbf{7.91} & 8.75  \\ 
        \hline
        \end{tabular}
    \end{center}
    \end{minipage}
    \hfill
    \begin{minipage}{0.32\textwidth}
    \begin{center}
        \footnotesize
        % \small
        \caption{Time Analysis (Unit: ms)}
        \label{tab:Efficiency}
        \tabcolsep=0.2cm
        \begin{tabular}{c|ccc}
        \hline
        \multirow{2}{*}{Method} & \multicolumn{3}{c}{Component}  \\
        & F. E. & Optim. &Track.  \\ 
        \hline
        LDSO~\cite{gao2018ldso} &- & -&20 \\
        ORB3~\cite{campos2021orb} &11 &126 &19  \\
        \hline
        Struct.~\cite{li2020structure} &38 &55 &42  \\
        PLP~\cite{shu2023structure} &48 &227 &55  \\
        VPL~\cite{chen2024vpl}  &21 &114 &32  \\
        Ours &21 &127 &33   \\ 
        \hline
        \end{tabular}
    \end{center}
    \end{minipage}%
\end{table*}

\subsection{Results}
\subsubsection{Results on Toy dataset}
We generated multiple sets of lines with approximate directions and continuous poses. Perturbations were added to the initial values of the poses and line observations involved in the optimization. As shown in Tab.\ref{tab:compression_time}, we compare the optimization time, the poses error, and the lines error with different representations (2-p~\cite{xu2022leveraging}, 4-p~\cite{bartoli2005structure}) of the line under three scenarios. In the scenario where the poses are fixed as ground truth and do not participate in the optimization, the line error in our method is slightly larger than the orthogonal representation, as the orthogonal representation fully utilizes an independent line. However, when perturbations are added to the poses and included in the optimization, our method's pose and line errors are smaller than those of the other two representations because our method integrates the structural regularities into BA. Fig.~\ref{fig:compareline} shows the convergence behavior of our method compared to the orthogonal representation~\cite{bartoli2005structure} under given perturbations. Furthermore, in all scenarios, our optimization speed is the fastest. This demonstrates that our method achieves a better trade-off between accuracy and speed. 
% \footnote{We modified the simulation data generation code referenced \href{https://github.com/HeYijia/vio_data_simulation}{here} and the comparison code from \cite{xu2022leveraging}.}

\subsubsection{Results on Indoor Datasets}
In structured indoor scenes characterized by low-texture regions and lighting variations, point-based SLAM systems perform poorly (especially methods based on optical flow), while line-feature systems provide structural information. Tab.\ref{tab:under} shows the tracking results for the indoor BeVIS~\cite{shao2023slam} and Garage~\cite{chen2024vpl} datasets. Our system achieved state-of-the-art tracking results in these low-texture and lighting-variable scenes. Additionally, we qualitatively visualized the generated point-line maps (Fig.\ref{fig:indoorVis}), where lines of different colors represent structural line features anchored to different principal axes. Our maps offer a clearer representation compared to other systems.

\subsubsection{Results on Outdoor Datasets}
We tested the tracking performance (Tab.\ref{tab:under} \ref{tab:kitti}) of our system on outdoor datasets KITTI~\cite{geiger2013vision} and Campus~\cite{chen2024vpl}. Following the classification in ~\cite{chen2024vpl}, we divided KiITTI's 11 sequences into structured scenes with structured buildings and semi-structured scenes with trees, poles, and other objects with less geometric information and constraints than buildings. In structured scenes (part of KITTI~\cite{geiger2013vision} and Campus~\cite{chen2024vpl}), our system effectively leveraged the scene's information, achieving good tracking metrics. In semi-structured scenes of KITTI~\cite{geiger2013vision}, our system can also effectively use the limited structural information, achieving relatively good results. It is worth noting that, apart from the direct method LDSO~\cite{gao2018ldso}, other feature-based methods failed in KITTI 01. However, as shown in the indoor dataset, LDSO~\cite{gao2018ldso} performs poorly in areas with lighting changes.

\subsection{Runtime Analysis}
Except for LDSO~\cite{gao2018ldso}, a direct method, all other methods are improvements on the ORB-SLAM~\cite{mur2017orb} series. Therefore, we tested these methods in terms of feature extraction (F.E.), local mapping thread optimization (Optim.), and the time required to track (Track.) a single frame (Tab.\ref{tab:Efficiency}). For LDSO, we only tested the time to track a single frame, as other modules cannot be fairly compared with ORB-SLAM~\cite{mur2017orb} series. Additionally, Structure-SLAM~\cite{li2020structure} is mainly designed to be suitable for indoor dataset, which reduces some optimized parameters during local optimization. The results show that our method consumes less time than line-based systems. This is primarily due to our principal axis anchoring optimization approach and the deliberate reduction in the number of point features that need to be detected.

\subsection{Ablation Study}
As Tab.\ref{tab:ablation2} shows, we conducted ablation experiments on principal axis update, data association algorithm, and structural preprocess. We test the impact with and without the update strategies for the principal axis. Regarding principal axis association, we performed ablation experiments by replacing our proposed soft association algorithm with hard association anchoring a line to only one principal axis. For the structural preprocess, we assessed the impact of having or not having the vertical prior and the effect of horizontal vanishing point refinement following the vertical prior. The results indicate that our strategies have been effective, especially the vertical prior and principal axis management. Additionally, our soft association strategy has somewhat improved accuracy, primarily because it reduces the weight of mismatched lines.

\begin{table}[t]
    \centering
    \caption{Ablation on Axis Management and Structural Preprocess}
    \tabcolsep=0.28cm
    \begin{tabular}{c|ccc|cc}
    \hline
    \multirow{2}{*}{Method} & \multicolumn{3}{c|}{Garage}  & \multicolumn{2}{c}{KiTTi} \\ 
    & s1 & s2 & s3 & 04 & 07 \\ 
    \hline
    Hard Association & 0.73 & 0.84 & 1.01 & 0.86  & 1.68\\
    w/o. Axis Management & 0.87 & 0.99 & 1.03 & 1.14 & 2.01 \\
    w/o. Vertical Prior  & 1.13 & 1.07 & 1.01 & 0.98 & 2.31 \\
    w/o. Vp Refinement & 0.69 & 0.81 & 0.96 & 0.73 & 1.73 \\
    Our Full Model  &0.69  &0.75  &0.91 & 0.72  & 1.52 \\
    \hline
    \end{tabular}
    \label{tab:ablation2}
\end{table}

\section{Conclusion}
Our approach successfully integrates the optimization of line features with the utilization of scene structural information in point-line SLAM systems. By anchoring co-directional lines to a principal axis and optimizing them with a reduced parameter set, we address the challenges of line structure representation and optimization. Our principal axis management algorithm, combined with vertical priors and vanishing points in the Atlanta World hypothesis, further enhances system robustness and minimizes mismatches.  In the future, we plan to extend our system to work with stereo, RGB-D, and IMU sensors, building a more robust and versatile SLAM system.
% The experimental results on multiple datasets and ablation studies validate the performance and reliability of our system. Considering the convenience of monocular setups, our system is currently designed for monocular applications.

% Additionally, we aim to incorporate deep learning methods for point and line feature detection and matching,
{
    \clearpage
    \bibliographystyle{IEEEtran}
    \bibliography{main}
}

\end{document}